\documentclass[conference]{IEEEtran}
\IEEEoverridecommandlockouts
\usepackage{multirow}
\usepackage{cite}
\usepackage{amsmath,amssymb,amsfonts}
\usepackage{algorithmic}
\usepackage{graphicx}
\usepackage{textcomp}
\usepackage{xcolor}
\usepackage{color, colortbl}
\definecolor{Gray}{gray}{0.9}
\definecolor{LightCyan}{rgb}{0.88,1,1}
\usepackage{booktabs}
\usepackage{enumitem}

\begin{document}

\title{EVOKE: Emotion Enabled Virtual Avatar Mapping Using Optimized Knowledge Distillation\\
\thanks{{\large$^{\ddagger}$}Authors with equal contributions.}
}


\author{
\IEEEauthorblockN{Maryam Nadeem\IEEEauthorrefmark{1},
Raza Imam\IEEEauthorrefmark{1}\IEEEauthorrefmark{3}, Rouqaiah Al-Refai\IEEEauthorrefmark{1}\IEEEauthorrefmark{3}, Meriem Chkir\IEEEauthorrefmark{1},\\ Mohamad Hoda\IEEEauthorrefmark{1}\IEEEauthorrefmark{2} and Abdulmotaleb El Saddik\IEEEauthorrefmark{1}\IEEEauthorrefmark{2} }
\IEEEauthorblockA{
\IEEEauthorrefmark{1}Mohamed Bin Zayed University of Artificial Intelligence (MBZUAI)\\
\IEEEauthorrefmark{2}University of Ottawa \\}
\textit{\{maryam.nadeem, raza.imam, rouqaiah.al-refai, meriem.chkir, mohamad.hoda, a.elsaddik\}@mbzuai.ac.ae}\\
\textit{elsaddik@uottawa.ca}
}

\maketitle

\begin{abstract}


As virtual environments continue to advance, the demand for immersive and emotionally engaging experiences has grown. Addressing this demand, we introduce \underline{E}motion enabled \underline{V}irtual avatar mapping using \underline{O}ptimized \underline{K}nowledg\underline{E} distillation (EVOKE), a lightweight emotion recognition framework designed for the seamless integration of emotion recognition into 3D avatars within virtual environments. Our approach leverages knowledge distillation involving multi-label classification on the publicly available DEAP dataset, which covers valence, arousal, and dominance as primary emotional classes. Remarkably, our distilled model, a CNN with only two convolutional layers and 18 times fewer parameters than the teacher model, achieves competitive results, boasting an accuracy of 87\% while demanding far less computational resources. This equilibrium between performance and deployability positions our framework as an ideal choice for virtual environment systems. Furthermore, the multi-label classification outcomes are utilized to map emotions onto custom-designed 3D avatars.

\end{abstract}

\begin{IEEEkeywords}
emotion recognition, EEG signals, 3D avatars, knowledge distillation, wellbeing
\end{IEEEkeywords}

\section{Introduction}

\begin{figure}[t]
    \centering
    \includegraphics[width=0.50\textwidth]{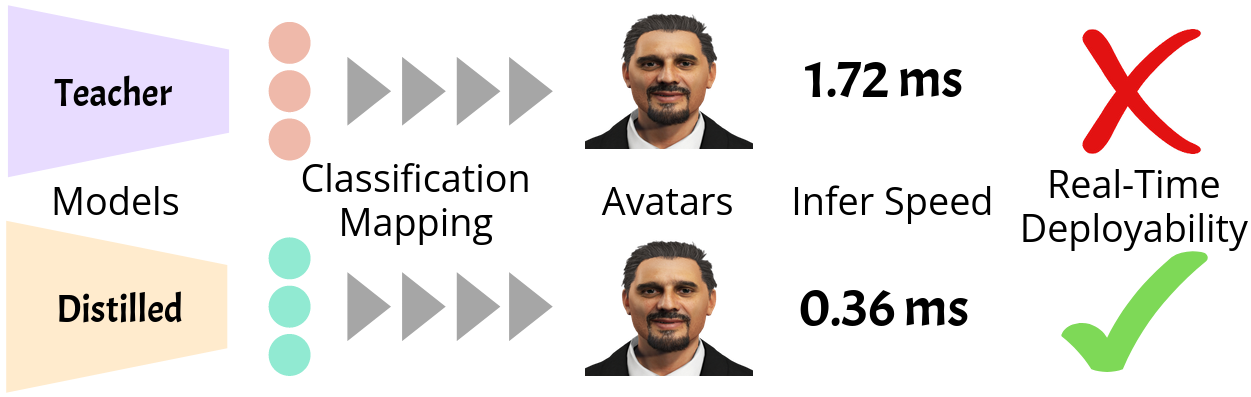}
    \caption{Inference speed of teacher and student models. Distilled student model can be deployed for real-time applications, as smaller models are faster with lower inference times.}
    \label{fig:TeachervsStudent}
\end{figure}

Nowadays, there is a growing need for better human-computer interactions, especially in expressing emotions in virtual environments \cite{cowie2001emotion}. Emotions can be recognized using text, speech, facial expressions, gestures, and physiological signals \cite{wang2022systematic}. Utilizing EEG signals for emotion recognition has evolved from defining certain sets of emotions to recognizing a wide range of them \cite{bos2006eeg, suhaimi2020eeg, liu2011real}. Mapping these emotions to avatar expressions in virtual environments can enhance the user experience and facilitate human-computer interactions in various applications such as healthcare and education, leading to advanced digital connectivity.
Different frameworks have been introduced for emotion recognition using EEG signals, ranging from traditional machine learning algorithms to more complicated deep learning approaches \cite{wang2014emotional, bazgir2018emotion}. However, deploying these approaches in real-time can be a challenging task because of the huge number of parameters, resource-constrained environments, and the computational power needed to run them. In applications where fast processing is needed, using a lightweight and accelerated approach becomes a priority. Our primary objective is to bridge the gap between emotion recognition in the world of digital avatars while focusing on their smooth integration into immersive virtual environments. To achieve this, we propose a knowledge distillation-based framework for multi-label classification that enables the transfer of features from a comparatively larger and more complex model (teacher model) to a smaller, more efficient one (student model). This distilled student model retains much of the teacher's accuracy but operates with significantly reduced computational resources \cite{imam2023seda}, making it highly suitable for deployment in resource-constrained environments or real-time applications as illustrated in Fig \ref{fig:TeachervsStudent}.

EVOKE finds meaningful application in the realm of healthcare and emotional well-being. It enables the development of recognition systems that can benefit patients and therapists by establishing a baseline of a patient's typical emotional states through extended EEG data collection, serving as a valuable health indicator. Moreover, visual representations of emotions via avatars enhance communication between therapists and patients, allowing patients to express their emotional experiences more effectively, thereby improving therapy outcomes. This approach also opens doors to virtual therapy and support groups within virtual environments, offering increased accessibility to emotional well-being, particularly for individuals with social anxiety or physical limitations.

The major contributions of this paper are as follows:
\begin{itemize}[label={--}]
    \item Proposed lightweight distilled model for EEG-based emotion recognition called EVOKE, which significantly reduces computational parameters by a factor of $18$x as compared to the originally trained model while maintaining a comparable performance.
    \item This work introduces a combination Binary Cross Entropy with Logits Loss and the concept of knowledge distillation for multi-label classification, which to our current understanding has not been explored previously for this task.
    \item Personalized mapping of the multi-label classification outputs to custom made 3D avatars ready to be deployed in any virtual environment. 
\end{itemize}

\section{Related Works}

\textbf{EEG and Emotion Recognition}.
Research on EEG-based emotion recognition algorithms has been growing rapidly over the past few decades. In 2009, a hierarchical binary classification approach was tested for EEG-based emotion classification \cite{lin2009eeg}. EEG emotion recognition problems have shown improved performance when utilizing deep learning techniques. For instance, in a recent work, Xiao et al \cite{xiao20224d} introduced a novel method, the four-dimensional attention-based neural network (4D-aNN). It transforms raw EEG signals into 4D spatial-spectral-temporal representations and uses attention mechanisms to assign weights to brain regions and frequency bands. Liu et al. \cite{liu20213dcann} introduced a three-dimensional convolutional attention neural network called 3DCANN which extracts dynamic relationships among multi-channel EEG signals over time and fuses spatio-temporal features with attention weights outperforming existing models. Yang et al. \cite{yang2018continuous} introduce a 3D representation of EEG segments to amalgamate features from various frequency bands while retaining spatial information among channels using a continuous CNN. Another research highlights attention mechanisms in EEG signal analysis using vision transformer-based methods \cite{arjun2021introducing}.  

\textbf{Emotion Recognition and Digital Avatars}.
There are two theoretical models for emotions, one is known as the discrete model which has a set of 6 basic emotions initially introduced by  \cite{ekman1969pan} and it was later expanded to 15 emotions. In contrast, there is the dimensional model that expresses a wide range of emotional states in two or three dimensions. In a multi-dimensional space, emotions are expressed with multiple fundamental features. According to Russell et al. \cite{russell1980circumplex}, emotions are mapped to two different dimensions, valence and arousal. While 2D model can span many emotions, it can struggle when the valence and arousal have the same levels. Therefore, to address this, Russell and Mehrabian \cite{russell1977evidence,mehrabian1996pleasure} introduced dominance as a third dimension for differentiation. These emotion schemes were used mainly for devoloping emotion recognition datasets but they can also be utilized for connecting the research with real world applications. For instance, in 2010, a fractal dimension-based algorithm for real-time EEG-based emotion recognition visualizing emotions through Haptek avatars was introduced \cite{liu2010real}. While this emotion mapping to digital avatars has its applications, they lack realism, expressiveness, and adaptability to virtual environment.

\begin{figure*}[h]
    \centering
    \includegraphics[width=0.7\textwidth]{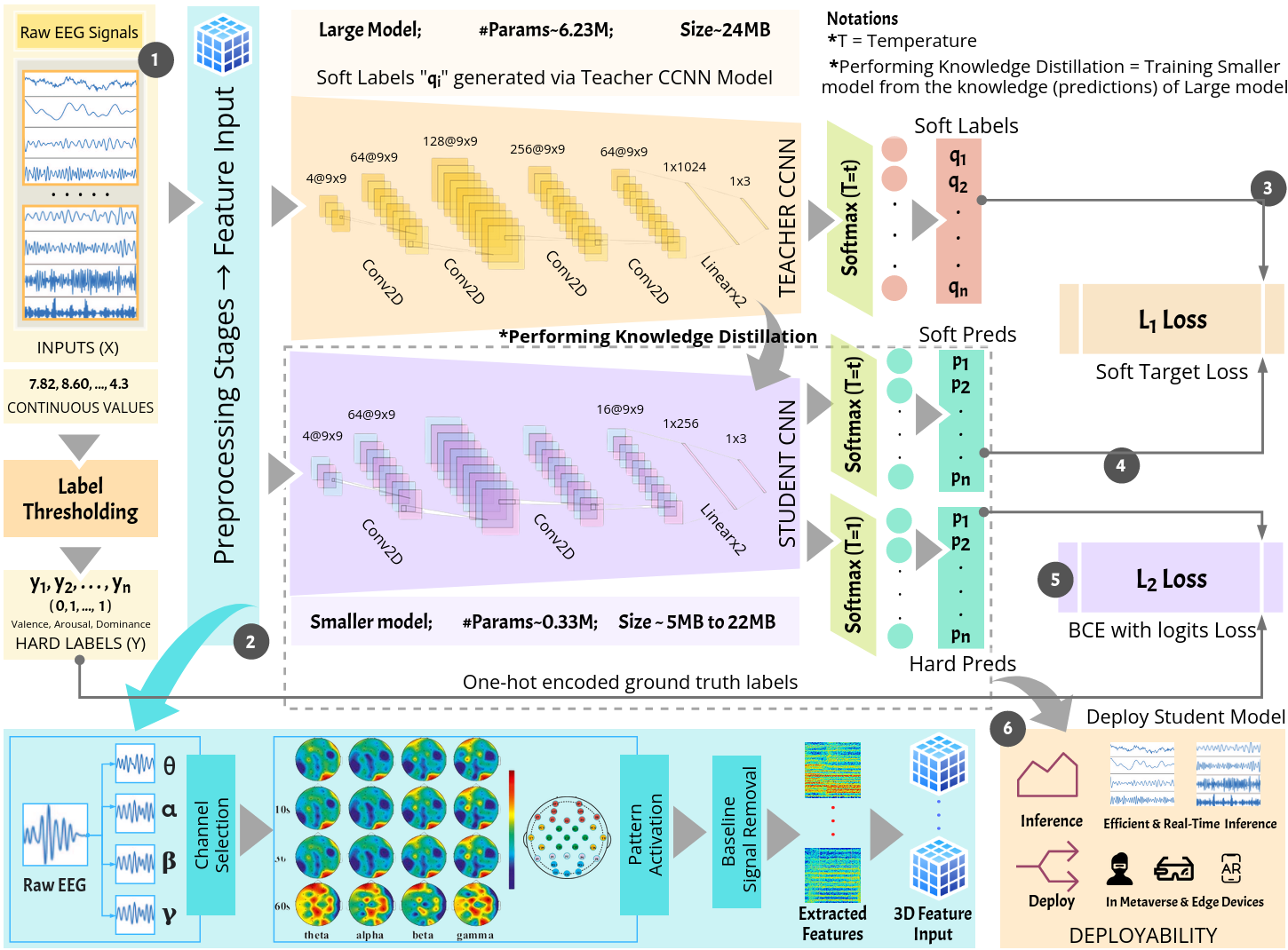}
    \caption{Our proposed framework, EVOKE. (1) We input the raw EEG signals to the framework which initially goes for preprocessing and label thresholding. (2) Preprocessing includes stages of feature extraction using differential entropy to have 4 channel bands followed by noise reduction resulting into 3D grid input. (3) Features input to teacher CCNN generates soft labels with Temperature $(T)>1$ which calculates the Binary Cross-Entropy with logits loss (\(\mathcal{L}_{1}\)). (4) Final distillation loss (\(\mathcal{L}_{distill}\)) is a weighted combination of soft target loss (\(\mathcal{L}_{1}\)) and final loss (\(\mathcal{L}_{2}\)) (refer Eq. \ref{kd_loss}). (5) The distilled model trained on the soft predictions of teacher model is then (6) deployed for fast inference and real-time applications.}
    \label{fig:method}
\end{figure*}

\section{Methodology}

\subsection{Preprocessing}
\subsubsection{Signal}
In the preprocessing of EEG signals, several steps were conducted to enhance data quality. Initially the original data was downsampled to 128Hz and artifacts from eye movements (EOG) were eliminated following established procedures \cite{koelstra2011deap}. Moreover, a bandpass frequency filter, spanning 4.0Hz to 45.0Hz, was applied to isolate relevant EEG frequencies. The data was subsequently averaged to establish a common reference baseline. To ensure consistent channel ordering, the EEG channels were rearranged according to the prescribed Geneva order. The data was then segmented into 60-second trials, with a preceding of 3-second pre-trial baseline removal (see Fig. \ref{fig:method} (2)). 
Furthermore, to enhance data coherence and relevance, the trial sequences were reorganized to align with experiments ID. In order to extract relevant features from raw EEG signals, differential entropy (DE) denoted as ($h(X)$) was employed to construct features in four frequency bands ($alpha\colon 8-14Hz$, $beta\colon 14-31Hz$, $gamma\colon 31-49Hz$, $theta\colon 4-8Hz$) \cite{duan2013differential}. Assuming a signal $X$ follows a probability Gaussian distribution $f(X) \sim \mathcal{N}(\mu, \sigma^2)$, then;

$$
    h(X) = -\int_{-\infty}^{\infty} f(X) \log (f(X)) dx
$$

\noindent where $ f(X) = (1/\sqrt{2\pi\sigma^2}) \exp (-(x-\mu)^2/2\sigma^2)$. Further, we removed the baseline signal and noise interference that was not associated with the emotional stimulus. To retain the spatial information and ensuring model compatibility, the signals were transformed into a grid-like 3D representation to be aligned with the positions of the electrodes.
\subsubsection{Label}
We applied a binary transformation to the emotional values using a consistent threshold of 5.0 for all categories (see Fig. \ref{fig:method} (1)), including valence, arousal, and dominance. This step streamlined the subsequent correlation into eight distinct emotions after the classification process is completed. Labels will be set to 1 if their value is larger than the threshold and set to 0 if it is smaller. This procedure was implemented to streamline the classification process.

\subsection{Knowledge-Distillation}
To develop a readily deployable model, we incorporated knowledge distillation into the training process. Following the foundational methodology introduced by Hinton et al. \cite{hinton2015distilling}, and making certain adaptations to the implementation for our specific task. 

Understanding the similarity between instances is crucial in comprehending the knowledge acquired by neural networks, especially in the context of multi-label classification of EEG signals into valence, arousal, and dominance. This is vital because gaining insights into how EEG patterns corresponds to these emotions interrelate can significantly enhance system accuracy.
In the standard sigmoid activation function, the output (\(\sigma(x)\)) is a \textit{hard} binary decision, mapping inputs to either 0 or 1. It's a step-like function with a sharp transition at \(x = 0\). 
To mitigate this, distillation incorporates a temperature parameter, denoted as $T$, into the sigmoid activation ($g$), the output is \textit{softened}.
$$g(x) = \begin{cases}
      0 & \text{if } x/T \to 0^+ \\
      1 & \text{if } x/T \to 0^-
   \end{cases}$$

When $T$ is set to 1, it corresponds to the standard sigmoid operation. This temperature-based operation is represented as:
$$q_i = \frac{1}{1 + \exp(-z_i / T)} \hspace{0.1cm} ,  \hspace{0.15cm}  p_i = \frac{1}{1 + \exp(-v_i / T)}$$

Here, \(z_i\) represents the logit for each class, and \(q_i\) transforms the teacher logits into probabilities, with $T$ controlling the level of smoothness.  During knowledge distillation, the teacher imparts its knowledge in the form of soft targets, calculated using this modified sigmoid with \(T>1\). If $v_i$ represents the logits by the student model then the student probabilities are represented as $p_i$.

The knowledge distillation process combines two distinct objective functions, \(\mathcal{L}_{1}\) and \(\mathcal{L}_{2}\). The first objective function, \(\mathcal{L}_{1}\), referred to as the soft target loss, captures the knowledge in the soft targets provided by the teacher model. In our case, it calculates the Binary Cross-Entropy (BCE) with logits loss between the teacher's soft targets (\(q_i\)) and the student's predictions (\(p_i\)), scaled by the square of the temperature (\(T\)). If $N$ denotes the number of samples in the dataset and $C$ denotes the number of emotions, then,
$$
\mathcal{L}_1 = -\frac{1}{N}\sum_{i=1}^{N}\sum_{j=1}^{C}\left[q_{i,j}\log(p_{i,j}) + (1 - q_{i,j})\log(1 - p_{i,j})\right] \cdot T^2
$$
The second objective function, \(\mathcal{L}_{2}\), aims to align the student's predictions (\(S_i\)) with the true labels (\(Y_i\)), while referencing the soft targets (\(q_i\)) from the teacher model,
$$
\mathcal{L}_2 = -\frac{1}{N}\sum_{i=1}^{N}\sum_{j=1}^{C}\left[Y_{i,j}\log(p_{i,j}) + (1 - Y_{i,j})\log(1 - p_{i,j})\right]
$$
The aim is to minimize the student model's distillation loss, \(\mathcal{L}_{distill}\), which combines \(\mathcal{L}_1\) and \(\mathcal{L}_2\) with a weighting factor \(\alpha\):
\begin{equation} \label{kd_loss}
\mathcal{L}_{distill}=\alpha * \mathcal{L}_{1} + (1 - \alpha) * \mathcal{L}_{2}   
\end{equation}

This combined loss ensures that the student model effectively captures the knowledge distilled from the teacher model while maintaining alignment with the true labels. Fig. \ref{fig:method} (3), (4), and (5) depict the process discussed above.

\begin{table*}[t]
\centering
\caption{Comparison of the experimented model properties in terms of performance and computational parameters.}
\label{tab:model_props_and_performance}
\renewcommand{\arraystretch}{1.2}
\resizebox{\textwidth}{!}{%
\begin{tabular}{l@{\hspace{10mm}} llllll@{\hspace{10mm}}ll @{\hspace{10mm}} ll}
\hline
& \multicolumn{6}{c}{Computational Performance} & \multicolumn{2}{c}{Differential Entropy} & \multicolumn{2}{c}{Power Spectral Density} \\ \cmidrule{2-7} \cmidrule{8-11} 
Model & \#Layers & \#Params & FLOPs & Weight & Inference (ms) & Throughput & F1 Score & Accuracy (\%) & F1 Score & Accuracy (\%)\\
\hline
ViT \cite{dosovitskiy2021image} & 140 & 85.64M & 16.86G & 321.56MB & 6.96 & 277.25 & 0.80 & 79.74 & 0.74 & 75.21 \\
Arjun ViT \cite{arjun2021introducing} & 41 & {144.356K} & {767.592K} & {587.35KB} & 1.6691 & 20110.31 & 0.65 & 62.24 & 0.77 & 77.78 \\
CCNN \cite{yang2018continuous} & 12 & 6.235M & 79.961M & 23.79MB & 0.6414 & 24572.37 & 0.92 & 93.23 & 0.46 & 61.11 \\
\rowcolor{LightCyan} 
\textbf{EVOKE (Ours)} & \textbf{8} & \textbf{353.363K} & \textbf{1.991M} & \textbf{1.35MB} & \textbf{0.3300} & \textbf{80176.24} & \textbf{0.88} & \textbf{87.62} & - & - \\
\hline
\end{tabular}
}
\end{table*}

\subsection{Model Architectures}
\subsubsection{Teacher}
We employed Continuous Convolutional Neural Network (CCNN) \cite{yang2018continuous} architecture as the teacher model for distillation, which consists of four convolutional layers with no pooling layers between them. The first three convolutional layers uses a $4$x$4$ kernel size and a stride of $1$. After each convolution operation, ReLU activation function was applied. To fuse different feature maps and reduce computational cost a $1$x$1$ convolutional layer with 64 feature maps was added. Following these four continuous convolutional layers, a fully connected layer to map the (64x9x9) feature maps into a final feature vector of size 1x1024 is added, followed by a final softmax layer for classification.

\subsubsection{Student}
The student model, a lightweight neural network, designed for multi-label classification of EEG signals into valence, arousal, and dominance, employs a 2 convolution layered architecture. The input, denoted as \(Z\), represents EEG signal data organized in a grid-like fashion with shape \([n, 4, 9, 9]\), where \(n\) signifies the batch size, \(4\) indicates the number of input channels corresponding to different electrodes, and \((9, 9)\) denotes the grid size. The model comprises two primary components: feature extraction and classification.
The feature extraction phase consists of two convolutional layers, \(c_1\) and \(c_2\), followed by Rectified Linear Unit (ReLU) activation functions. These layers process the input EEG data, producing intermediate outputs \(Z^{(1)}\) and \(Z^{(2)}\), respectively. 
The flattening operation then transforms \(Z^{(2)}\) into a one-dimensional tensor \(Z^{(3)}\). Let $G$ represent the classifier or the MLP, then, the feature representation, denoted as $Z^{(3)}$ is passed through $G$ to predict the output labels:
\begin{align*}
\hat{y}_{\text{valence}} &= G(Z^{(3)}) \in \{0, 1\} \\
\hat{y}_{\text{arousal}} &= G(Z^{(3)}) \in \{0, 1\} \\
\hat{y}_{\text{dominance}} &= G(Z^{(3)}) \in \{0, 1\}
\end{align*}
In our specific case, this classifier enables multi-label classification for the 3 umbrella classes, namely valence (\(y_{\text{valence}}\)), arousal (\(y_{\text{arousal}}\)), and dominance (\(y_{\text{dominance}}\)).

\begin{figure}[t]
    \centering
    \includegraphics[width=8cm]{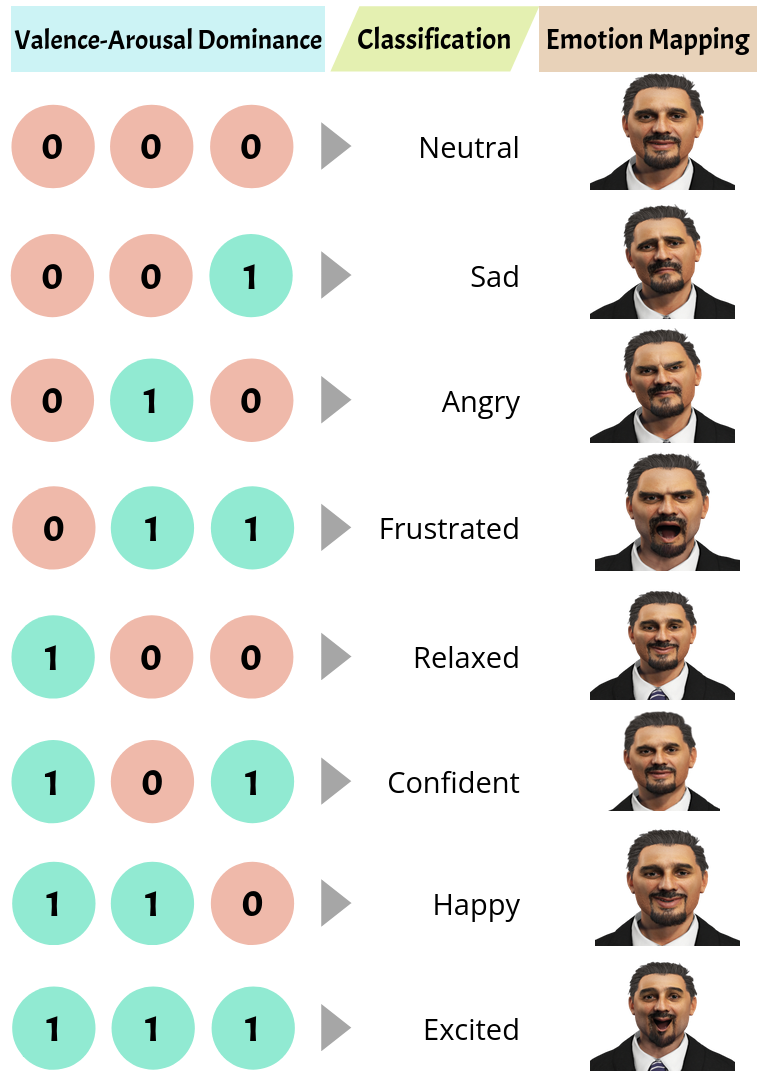}
    \caption{Multi-label classification results are mapped to eight emotions based on different combinations of valence, arousal, and dominance and further associated with 3D avatars through a hashing process.}
    \label{fig:mapping}
\end{figure}

\section{Experiments and Results}
\subsection{Dataset}
The DEAP dataset, developed by Koelstra et al. \cite{koelstra2011deap}, was employed for our study. It comprises EEG signals from 32 subjects watching a series of 40 one-minute music videos, each accompanied by emotional response ratings on scales of arousal, valence, dominance, liking, and familiarity. In our study, we focus exclusively on arousal, valence, and dominance, as liking and familiarity are more related to individual perspectives \cite{koelstra2011deap}. DEAP's 32-channel EEG data collection during emotional stimuli distinguishes it in EEG emotion recognition research, offering richer features for improved emotion state differentiation. Consequently, it serves as the exclusive dataset in our research.

 \begin{figure*}[h]
\centering
\begin{minipage}{.24\textwidth}
  \centering
  \includegraphics[width=\textwidth]{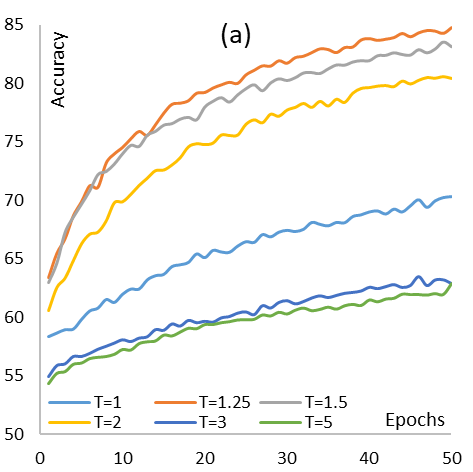}
\end{minipage}%
\begin{minipage}{.24\textwidth}
  \centering
  \includegraphics[width=\textwidth]{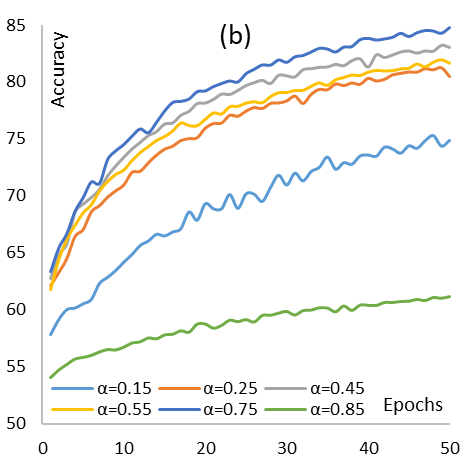}
\end{minipage}
\begin{minipage}{.24\textwidth}
  \centering
  \includegraphics[width=\textwidth]{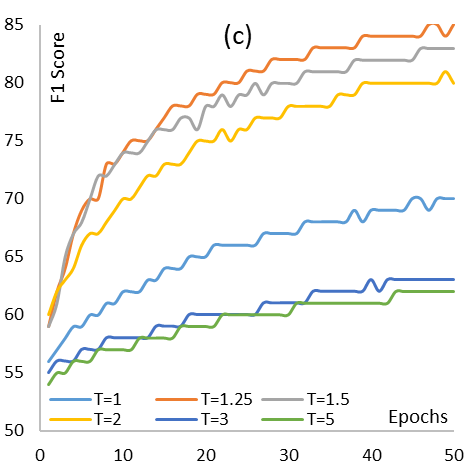}
\end{minipage}%
\begin{minipage}{.24\textwidth}
  \centering
  \includegraphics[width=\textwidth]{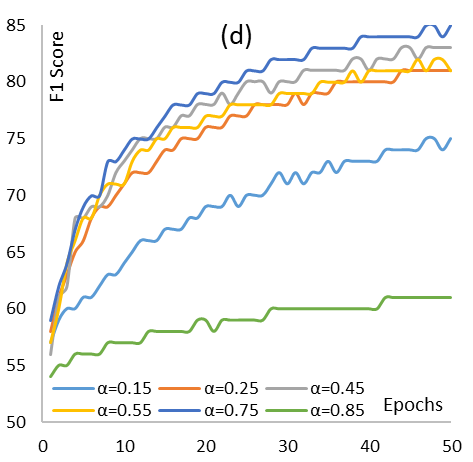}
\end{minipage}
\caption{Performance analysis in terms of accuracy and F1 score, respectively, across various values of temperature ($T$) (a) and (c) parameter and weight factor $\alpha$ (see Eq. \ref{kd_loss}) (b) and (d). Note that the accuracy and F1 scores presented in the figures are based on the mean values obtained from a 5-fold cross-validation evaluation.}
\label{fig:performance}
\end{figure*}

\subsection{Implementation Details}
The optimization process employed the Adam optimizer with a learning rate set to \(1 \times 10^{-3}\). ReLU activation layers were used for the student model. A consistent batch size of \(128\) was applied across all models. The training phase extended for \(100\) epochs and encompassed a 5-fold cross-validation strategy. The entire experimentation process was conducted within the PyTorch framework \cite{paszke2019pytorch}, utilizing a single Nvidia RTX A6000 GPU with 40 GB of memory.

For creating 3D avatars, we used Character Creator 4's \footnote{https://www.reallusion.com/character-creator/} Headshot plugin to turn a character's image into a 3D face mask. This involved adjusting expressions, hair color, and skin tone. Then, in Blender\cite{mullen2011mastering}, we created clothing based on the 3D avatar. Importing it back to Character Creator ensured proper rigging. After fine-tuning, we had a lifelike 3D avatar, blending real-world aesthetics with virtual artistry. Headshot, a key feature of Character Creator, analyzed real-life photos to generate a detailed 3D face, accounting for contours, skin texture, and expressions.

\subsection{Model Selection and Pretraining}
We conducted a meticulous evaluation of three state-of-the-art Emotion-Recognition models: Arjun ViT \cite{arjun2021introducing}, ViT \cite{dosovitskiy2021image}, and CCNN \cite{yang2018continuous}. We started with a clean slate, training these models entirely from scratch without the application of any knowledge distillation techniques. Our objective was to select the optimal candidate from these pretrained models to serve as the teacher model for our framework. For model evaluation, we employed \textit{accuracy} and \textit{F1-score}, which offer comprehensive insights into a model's performance in predicting multiple labels \cite{benedict2021sigmoidf1}. These metrics were derived from the mean values obtained through five-fold cross-validation.

The models were first evaluated under two distinct EEG-based feature extraction techniques: Differential Entropy (DE) and Power Spectral Density (PSD) as presented in Table \ref{tab:model_props_and_performance}. Notably, the results demonstrate the superior performance of CCNN across both the evaluation metrics. This can be attributed to several key factors;
firstly, the transformer models, although meticulously customized for EEG data, are inherently data-hungry and struggled to harness the full potential of the dataset and exhibited comparatively limited generalization capabilities.
Secondly, the unique nature of EEG data, which includes spatial information pertaining to the arrangement of electrodes on the scalp, presents a significant advantage for CCNN. This spatial awareness empowers CCNN to extract fine-grained details from the EEG data, effectively capturing the nuances associated with emotional states. All subsequent experiments were conducted with CCNN as the teacher. For student model, we experimented various architectures for our custom CNN model, discovering that an 8-layer design with two convolutional layers was the most suitable.

\subsection{Computation and Performance}
 Our distilled model, EVOKE, stands out with significantly fewer parameters (18x lesser than the teacher model) while achieving the fastest inference time of $0.33$ ms and highest throughput of $80176$ compared to other models (see Table \ref{tab:model_props_and_performance}). It's performance surpasses that of ViT and ArjunViT, comparable with the performance of the teacher model. This compelling balance between performance and deployability makes this framework suitable for virtual environment systems. Note that we omitted the Power Spectral Density (PSD) feature extraction technique during EVOKE's training due to inferior results compared to those of differential entropy as shown in Table \ref{tab:model_props_and_performance}.
 Additionally, we conducted experiments involving various temperature parameter $(T)$ values, as shown in Figure \ref{fig:performance} (a) and (c). Our findings indicate that the model performs optimally when $T$ is set to $1.25$. Notably, performance starts to decline gradually for $T$ values exceeding $2$. Further, we also explored different values of $\alpha$, the weight factor in Eq. \ref{kd_loss}. The model achieved its best performance when $\alpha$ was set to $0.25$ (see Fig \ref{fig:performance} (b) and (d)). It was observed that excessively large or small $\alpha$ values resulted in a decreased performance. In Fig. \ref{fig:performance}, the plots illustrate the model's performance during the initial $50$ epochs.

\subsection{Emotion Mapping to 3D Avatars}
The three-dimensional emotions model namely Valence-Arousal-Dominance or VAD model for short, includes the basic emotions \cite{ekman1969pan} defined by the rating of each dimension. For instance, the closer the rating of each dimension to zero  the lower the emotion distinction. Which means as shown in Fig. \ref{fig:mapping} when all categories are low or zero then the emotion is neutral.  Hence, in this paper, we have developed a set of combinations and their corresponding emotion mappings that bridge the gap between emotion classification and its representation in 3D avatars. Unlike some prior approaches \cite{liu2010real}, which utilized the 2D Valence-Arousal emotion model, our work incorporates an additional dimension of dominance. This additional dimension allows for a broader range of emotions, including neutral and excited states, alongside the six fundamental emotions. The number of emotions is limited intentionally to maintain focus and manageability, especially concerning the emotion mapping onto avatars. The combinations and their emotion mapping are shown in Fig. \ref{fig:mapping} where 0 indicates the low level of the primary emotion and 1 indicate having a high level for that emotion.


\section{Conclusion and Future Work}

In response to the practical deployment of real-time EEG emotion classification, we introduced EVOKE, a knowledge distillation-based lightweight model for emotion recognition and integration into virtual environments. The framework maps multi-label classification results to eight distinct emotions and links them to custom-created 3D avatars. When tested on the publicly available DEAP EEG dataset, our proposed model achieved competitive performance along with significantly fewer parameters as compared to other state-of-the-art models. This work paves the way for enhanced emotional communication in virtual settings, offering numerous applications in healthcare, therapy, and beyond, ultimately making emotional well-being more accessible and immersive.

One notable application is in Virtual Therapy Sessions:
\begin{itemize}[label={--}]
    \item \textbf{Scenario:} In virtual therapy sessions, EVOKE can be utilized to enhance the emotional interaction between therapists and clients by mapping real-time emotional states onto virtual avatars. 
    \item  \textbf{Application:} As clients express their feelings and emotions, EVOKE processes the emotional cues through its knowledge-distilled model, providing therapists with a visual representation of emotional changes in the virtual avatars. This visual feedback aids therapists in understanding and responding empathetically to the client's emotional state.
\end{itemize}

In the future, we aim to integrate real-time EEG signals into a healthcare virtual environment system, making it readily accessible for use by healthcare professionals.
    

\bibliographystyle{unsrt} 
\bibliography{references} 

\end{document}